\documentclass[10pt,twocolumn,letterpaper]{article}

\usepackage{wacv}
\usepackage{times}
\usepackage{epsfig}
\usepackage{graphicx}
\usepackage{amsmath}
\usepackage{amssymb}
\usepackage{amsthm}
\usepackage{enumerate}
\usepackage{graphics}
\usepackage{multirow}
\usepackage{array}
\usepackage{booktabs}

\theoremstyle{definition}
\newtheorem{definition}{Definition}  

\newtheorem{property}{Property}

\usepackage{algorithm}
\usepackage[noend]{algpseudocode}

\makeatletter
\def\BState{\State\hskip-\ALG@thistlm}
\makeatother

\wacvfinalcopy 

\def\wacvPaperID{528} 

\ifwacvfinal\fi
\begin{document}

\title{ Prototypicality effects in global semantic description of objects}

\author{Omar Vidal Pino \hspace{2cm} \hspace{2cm} Erickson R. Nascimento \hspace{2cm}  Mario F. M. Campos \\
Universidade Federal de Minas Gerais (UFMG), Brazil\\
{\tt\small \{ovidalp, erickson, mario\}@dcc.ufmg.br}}

\maketitle
\ifwacvfinal\thispagestyle{empty}\fi

\begin{abstract}
In this paper, we introduce a novel approach for semantic description of object features based on the prototypicality effects of the Prototype Theory. Our prototype-based description model encodes and stores the semantic meaning of an object, while describing its features using the semantic prototype computed by CNN-classifications models. Our method uses semantic prototypes to create discriminative descriptor signatures that describe an object highlighting its most distinctive features within the category. Our experiments show that: i) our descriptor preserves the semantic information used by the CNN-models in classification tasks; ii) our distance metric can be used as the object's typicality score; iii) our descriptor signatures are semantically interpretable and enables the simulation of the prototypical organization of objects within a category.


\end{abstract}
%
%


\section{Introduction}

The extraction of image relevant features has been the subject of Computer Vision research for decades. The advent of Convolutional Neural Networks~(CNN) enabled to achieve a visual recognition model with similar behavior of \textit{semantic memory}~\cite{tulving2007coding} for classification tasks~\cite{he2016deep,simonyan2014very,szegedy2017inception}, and sparked the tendency of semantic processing of images using deep-learning techniques. 
 
For several years, hand-crafted features~\cite{bay2008SURF,lowe2004SIFT,tola2008DAISY} and machine learning~\cite{perez2013genetic,simonyan2014learning,strecha2012ldahash} were the choice methods for image feature description tasks. The impressive success of CNN-models spawned numerous CNN-descriptors produced by different approaches that learn effective representations for describing image features~\cite{han2015matchnet,kim2017fcss,simo2015discriminative,zagoruyko2015learning,zbontar2015computing}. 
Consequently, representations of image features extracted using deep classification models~\cite{he2016deep,simonyan2014very,szegedy2017inception}, or using  CNN-descriptors are commonly referred to as \textit{semantic feature} or \textit{semantic signature}.

The term \textit{semantic feature} has been extensively studied in the field of linguistic semantics and it is defined as the representation of the basic conceptual components of the meaning of any lexical item~\cite{fromkin2018introduction}. In the  seminal work of Rosch~\cite{rosch1975cognitive} the author analyzed the semantic structure of the meaning of words and introduced the concept of prototype semantics~(or Prototype Theory). According to Rosch~\cite{rosch1975cognitive,rosch1975family}, the representation of a category semantic meaning is related to the category prototype, particularly to those categories naming natural objects.

\begin{figure}[t]
	\begin{center}
		
		
		\includegraphics[width=0.9 \linewidth]{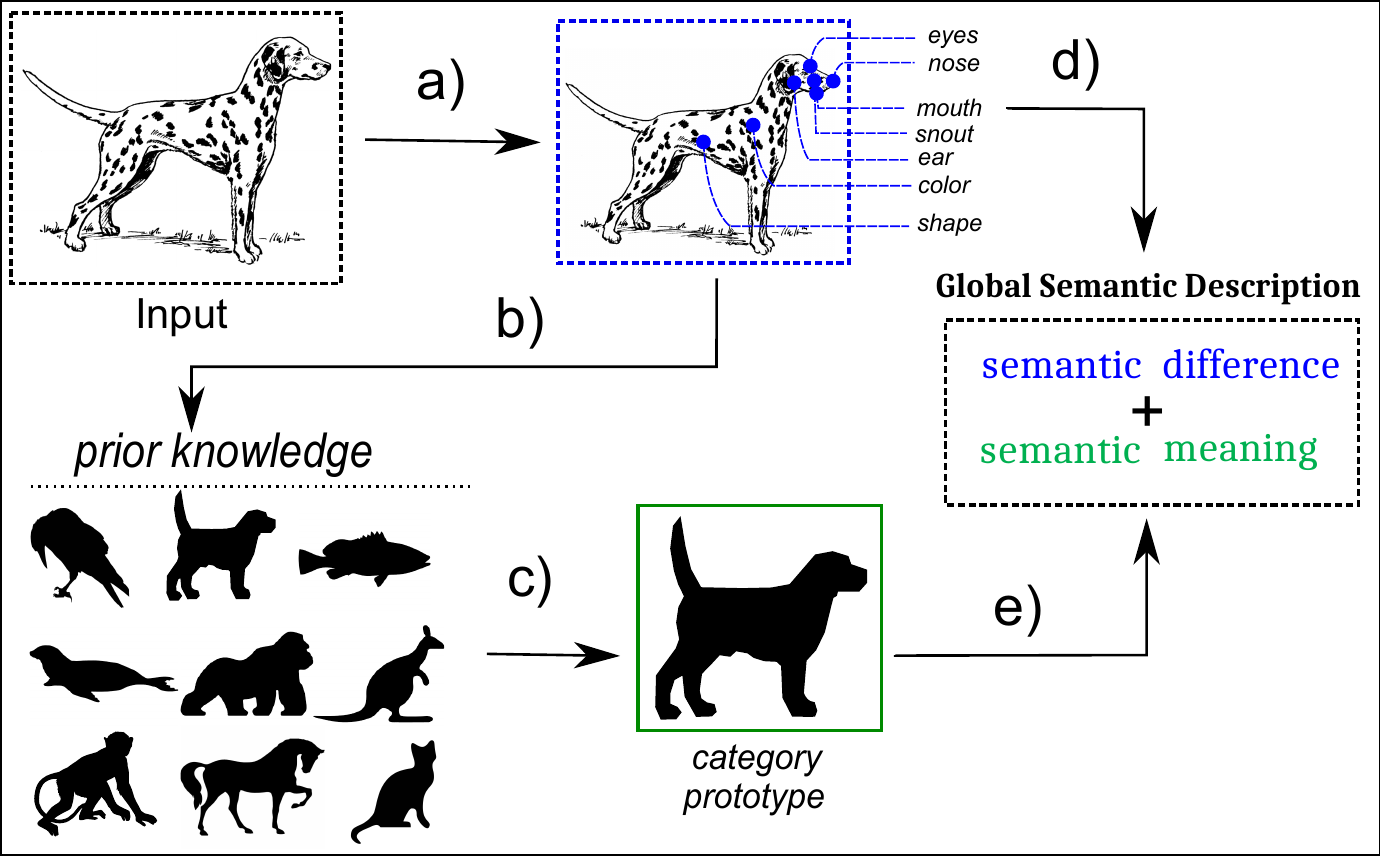}
	\end{center}
	\caption{\emph{Motivation and Concepts. Schematic of prototype-based description model}.
		\textit{a)} features extraction; \textit{b)} object features recognition; \textit{c)} categorization; \textit{d)} object features; \textit{e)} central semantic meaning of a category. The human visual system is able to observe an object and to build a global semantic description highlighting the object features that make it distinctive within the category. We propose how to simulate this behavior through the processing flow from  \textit{a)} to \textit{e)}.}
	\label{fig:motivation}
\end{figure}

Some CNN-description models~\cite{han2015matchnet,lin2016learning,simo2015discriminative,zagoruyko2015learning,zbontar2015computing}~(and semantic description models~\cite{choy2016universal,han2017scnet,kim2017fcss,Rocco18}) stand for the semantic information of the image features using a range of different approaches. Nevertheless, none of these models construct their representations coding the visual semantic information with the extensive theoretical foundation of Cognitive Science to represent the semantic meaning. We rely on cognitive semantics studies related to the Prototype Theory for modeling the central semantic meaning of category.
Our approach uses the representation of central semantic meaning of category for simulating the human behavior in object features description task.

In this paper, we propose a novel approach to take on the semantic features descriptions of objects. We bring to light the Prototype Theory as a theoretical basis to represent the semantic meaning of the image visual information. We develop a prototype-based description model that uses the category's prototype to find a global semantic representation of the basic conceptual components (objects) of the image semantic meaning.
Human beings can learn the most distinctive features of a specific category~\cite{martin2007representation,thompson2003neuroimaging}. These learned features (or properties) are used by the human brain to identify, classify and describe objects~\cite{tulving2007coding}. The Prototype Theory proposes that human beings think of a category in terms of abstract prototypes, defined by the central case of a category~\cite{geeraerts2010theories,rosch1975cognitive,rosch1978principles}. 

Successful execution  of the object recognition and description tasks in the human brain is inherently related to the learned prototype of the object category~\cite{minda2002comparing,rosch1975cognitive,rosch1978principles,zaki2003prototype}. This raises the following two questions: i) Can a model of the perception system be developed in which objects are described using the same semantic features that are learned to identify and classify them? ii) How can the category prototype be included in the object global semantic description?
We address these two questions motivated by the human approach for describing objects highlighting their most distinctive features within the category. For example, a typical human description: a dalmatian is a dog (semantic meaning) that is distinguished by its unique black or liver colored spotted coat~(semantic difference with respect to the central semantic meaning of dog category). Figure~\ref{fig:motivation} depicts our prototype-based description model.

We evaluate our approach using CNN-models both on MNIST and ImageNet datasets. The experiments show that our prototype-based description model can simulate the prototypical organization of objects categories. Furthermore, our descriptor can construct semantic signatures that are discriminative, interpretable, with low dimensionality, and with the ability to encapsulate and to retain the meaning of object features.



\section{Related works}

\paragraph*{CNN descriptors.} The CNN descriptor family showed that it is possible for a learning approach to outperform the best techniques based on carefully hand-crafted features~\cite{bay2008SURF,lowe2004SIFT,tola2008DAISY}. These models differ among themselves on how to compute the descriptors in their deep architectures, similarity functions and  features extraction methods. Some approaches extract immediate activations of the model as a descriptor signature~\cite{donahue2014decaf,fischer2014descriptor,gong2014multi,long2014convnets}. Other methods directly learned a measure of similarity to compare image patches using a similarity convolutional network~\cite{han2015matchnet,simo2015discriminative,yi2016lift,zagoruyko2015learning}. Siamese networks were used to learn discriminative representation and to learn a similarity metric~\cite{han2015matchnet,zagoruyko2015learning,zbontar2015computing}. The deep model LIFT~\cite{yi2016lift} learns each of the tasks involved in feature management: detection, orientation estimation, and feature description. Lin~\etal~\cite{lin2016learning} constructed a compact binary descriptor for efficient object matching based on the features extracted with the VGG16 model~\cite{simonyan2014very}.
\paragraph*{Semantic descriptors and correspondence.}  Finding correspondences between different scenes that share similar or semantically related features is a challenging problem. Liu~\etal~\cite{liu2011sift} propose to use SIFT Flow to create semantic flow family methods as a solution to the high degree of variation that includes the challenge of semantic correspondence~\cite{bristow2015dense,kim2013deformable,liu2011sift,qiu2014scale,trulls2013dense,yang2014daisy}. Several of these methods combine their approaches with the extraction of hand-crafted features~\cite{lowe2004SIFT,tola2008DAISY}. Some works~\cite{choy2016universal,han2017scnet,zhou2016learning} use the robustness of CNN-models for training deep learning architectures that  address the problem of semantic correspondence. Kim~\etal~\cite{kim2017fcss} tackled the problem of semantic correspondence by constructing a semantic descriptor. FCSS descriptor~\cite{kim2017fcss} has the property of being robust to intra-class appearance variation due of its local self-similarity (LSS) and its ability to keep the precise localization of deep neural networks.
The performance of CNN-models used in description tasks are still not at par with the performance achieved by CNN used in classification models. In general, CNN descriptors and semantic descriptors are trained to learn their own semantic representations and use different deep learning architectures. Most of these feature description models do not use the discriminative power of the features extracted using the well-know CNN-classification models~\cite{he2016deep,simonyan2014very,szegedy2017inception}. Moreover,  none of these feature description approaches incorporates the cognitive sciences foundation to introduce meaning in the representations of image features.



\paragraph*{Prototype Theory.} The Prototype Theory analyzes the internal structure of categories and introduces the prototype-based concept of categorization. It proposes a categories representation as heterogeneous and not discrete, where the features and category members do not have the same relevance within the category.  Rosch~\cite{rosch1975cognitive} obtained evidence that humans store the \textit{semantic meaning of category} based on the degrees of representativeness~(\textit{typicity}) of category members. The author showed that human beings store the category knowledge as a semantic organization around of category prototype~(\textit{prototypical organization})~\cite{rosch1978principles}.
The \textit{prototype} or \textit{prototypical concept} was formally defined as the clear central member of a category~\cite{geeraerts2010theories,rosch1975cognitive}. 
Rosch~\cite{rosch1978principles} showed that human beings learn first the core semantic meaning of the object~(\textit{prototype}) and then its specificities. 
In this paper, we model the central semantic meaning of category based on the four types of \textit{prototypicality effects}~\cite{geeraerts1997diachronic,geeraerts2010theories}: \textit{extensional non-equality}, \textit{intensional non-equality}, \textit{extensional non-discreteness}, and \textit{intensional non-discreteness}. The prototypicality effects surmise the importance of the distinction between central  and peripheral meaning~\cite{geeraerts2010theories}.

\section{Methodology}
Rosch~\cite{rosch1975cognitive, rosch1978principles} showed that humans learn the central semantic meaning of categories~(the prototype) and include it in their cognitive processes. Based in these assumptions, our proposal follows the flow of conceptual processes presented in Figure~\ref{fig:motivation} as hypothesis for simulating the human behavior in object features description. We propose to describe an object, highlighting the global features that distinguish it within a category. In other words, after recognizing the category to which the object belongs, how do we  find what are the features that distinguish it from others within the category? How to model a global object description with similar behavior of the diagram in Figure~\ref{fig:motivation}?

\begin{figure*}[t!]
	\begin{center}
		\includegraphics[scale=0.5,width=0.98\linewidth]{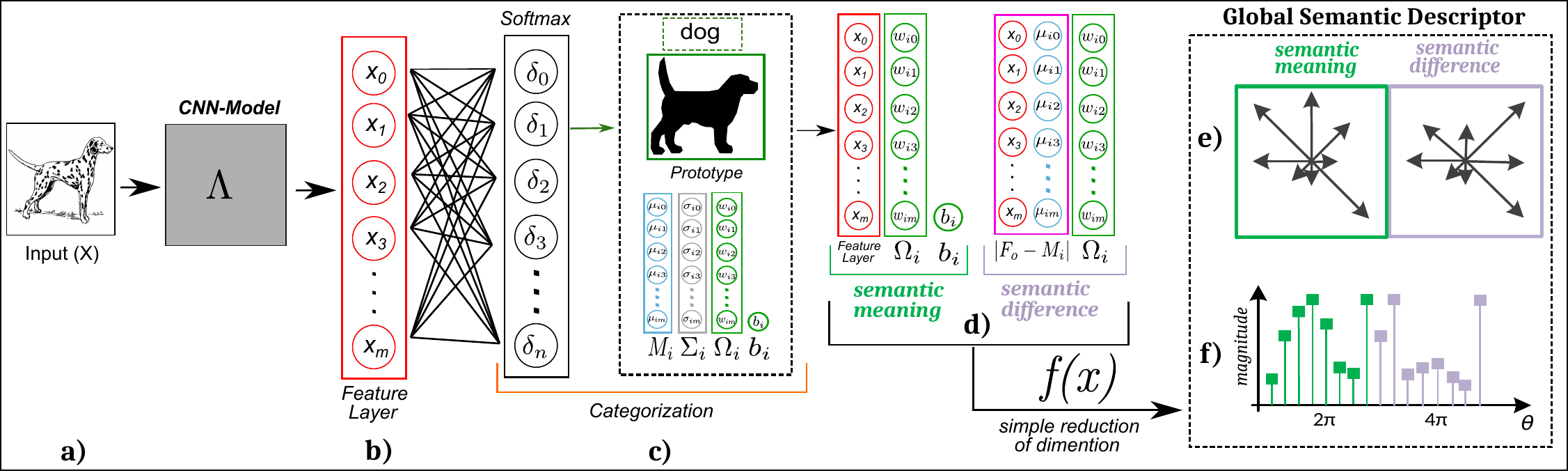}	
	\end{center}
	\caption{\textit{Overview of prototype-based description model.} Set of steps to transform the visual information received as input into a Global Semantic Descriptor signature.~\textit{a}) input image;~\textit{b)} extracted features using a CNN-classification model;~\textit{c)} classification and category prototype selection;~\textit{d)} Global semantic description of object using the category prototype;~\textit{e)} graphic representation of the Global Semantic Descriptor signature resulting from the dimensionality reduction function~($f(x)$); and ~\textit{f)} Global Semantic Descriptor signature.}
	\label{fig:methodology}
\end{figure*}

To address these issues, an due to their good performance, we use CNN-classification models in feature extraction, recognition and classification of the visual information received as input~(processes a, b, c, and d in Figure~\ref{fig:motivation}). The CNN-models, analogous to the human memory~\cite{fuster1997network}, make associations that keep the knowledge in their connection structures. Our method downloads that knowledge of pre-trained CNN-models into a semantic structure (semantic prototype) that stands for the central semantic meaning of learned categories~(Figure~\ref{fig:motivation}e)). Our method proposes a representation (signature) that describes an object, encapsulating the \textit{semantic meaning} of extracted features, and its \textit{semantic differences} in relation to the \textit{central semantic meaning} of the category.
In the following sections, we present part of our method that encapsulates the category central meaning (prototype). Also, we present how to introduce the prototype representation in semantic description of object features. Figure~\ref{fig:methodology} shows the architecture overview of our prototype-based description model.

\subsection{Prototype Construction}
\label{sec:prototype}

The semantic structure, \emph{i.e.}, \textit{central/peripheral meaning}, of a category are related with differences of typicality and membership salience of category members~(\textit{extensional non-equality}).
The prototype is an ``average'' of the abstraction of all objects in the category~\cite{sternberg2016cognitive}. It summarizes the most representative members~(or features) of the category. The combination of the observed features and their relevance for the category enables the grouping of objects into family resemblance~ (\textit{intensional non-equality}). This approach justifies the object's position within the semantic structure of the category and allows typical objects to be grouped into the semantic center of the category (\textit{prototypical organization}).

Let  $ C = \left\{  {c_1, c_2,...,c_n}\right\} $ be a finite set of categories of objects, $F =\left\{ {f_1, f_2,...,f_m}\right\} $ be a finite set of distinguishing  features of an object, and $O_{c_i} = \left\{{o \in O: category(o) = c_i}\right\}$, is the set of objects that share the same category ${c_i},$ (where $ O =\left\{{an\: universe\: of\: objects}\right\}$).


\theoremstyle{definition}
\begin{definition}{\textit{Semantic prototype.}}
	We call the \textit{central meaning} of the category  $c_i \in C$, \textit{semantic prototype} of $c_i$-category or simply \textit{semantic prototype} to the ``average'' and  standard deviation of each of the features of all objects within the category, along with a ``measure'' of the relevance of those features. Formally the semantic prototype is a $3$-\textit{tuple} $ P_i = \left( {M_{i},\Sigma_{i},\Omega_{i}}\right) $ where $\:  \forall i = 1,...,n;\forall j = 1,...,m $:  i) $ M_i = \left[  {\mu_{i1}, \mu_{i2},...,\mu_{im}}\right] $ is a nonempty $m$-dimensional vector, where $\mu_{ij}$ is the \textit{mean} of the j-$th$ feature extracted for \textit{only typical objects} of the category $c_i \in C$; ii) $ \Sigma_{i} = \left[ {\sigma_{i1}, \sigma_{i2},...,\sigma_{im}}\right] $ is a nonempty $m$-dimensional vector, where $\sigma_{ij}$ is the standard deviation of the j-$th$ feature extracted for \textit{only typical objects} of the category $c_i \in C$; iii) $ \Omega_{i} = \left[ {\omega_{i1}, \omega_{i2},...,\omega_{im}}\right] $ is a nonempty $m$-dimensional vector, where $\omega_{ij}$ is the relevance value of the j-$th$ feature for the category $c_i \in C$.
	%
	\label{def:semantic_pttype}
\end{definition}

\begin{definition}{\textit{Convolutional semantic prototype.}}
	The \textit{convolutional semantic prototype} of a category $c_i \in C$ is a $4$-tuple $ P_i = \left( {M_{i},\Sigma_{i},\Omega_{i},b_{i}}\right),$ where $M_{i},\Sigma_{i}$ are computed using features of category $c_i$ extracted from the \textit{fully convolutional layer} of CNN-models; and $\Omega_{i},b_{i}$ are the learned parameters of \textit{i}-$th$ category in the softmax layer. Next, we refer to the \textit{convolutional semantic prototype} of the category as \textit{semantic prototype}.
\end{definition}

%
%


\begin{definition}{\textit{Semantic value.}} 
	The \textit{semantic meaning} of observed features $F =\left\{  {f_{1}, f_{2},...,f_{m}}\right\}$ for category $c_i \in C$, \textit{summary value} of the observed features $F$, or simply \textit{semantic value} of $F$ in $c_i$-category is an abstract value:
	$\hat{z} = \sum_{m} \omega_{ij}a_j + b_{i},$ 
	where $\omega_{ij} \in \Omega_{i},\,a_{j} \in \left\{F, M_{i}\right\} $. Consequently, the \textit{central semantic meaning of the category} $c_i \in C$ or \textit{summary value} of the semantic prototype $P_i= \left( {M_{i},\Sigma_{i},\Omega_{i},b_{i}}\right)$ is the  \textit{semantic value} $\hat{z_{i}} = \sum_{m} \omega_{ij}\mu_j + b_{i},$ where $\omega_{ij} \in \Omega_{i},\,\mu_{ij} \in M_{i},\,\forall j = 1,...,m ;\,\forall i= 1,...,n.$ 
	\label{def:semantic_value}   
\end{definition}



\begin{definition}{\textit{Prototypical distance.}}
	Let $ {\emph{o} \in O_{c_i}}$ a representative object of category $c_i \in C$, $\emph{F}_o$ the features of object $\emph{o}$ and  $ P_i = \left( {M_{i},\Sigma_{i},\Omega_{i},b_{i}}\right)$ the semantic prototype of the category $c_i$. We defined as \textit{prototypical distance}  between $\emph{o}$ and $P_i$ the semantic distance:
	\begin{eqnarray} 
	\delta(\emph{o},P_i) = \sum_{j=1}^{m} \left| \omega_{ij}\right|\left|f_j-\mu_{ij}\right|
	\label{eq:pttype_distance}
	\end{eqnarray} 
	where $\omega_{ij} \in \Omega_{i},\,\mu_{ij} \in M_{i}, \,$ and $f_j \in F_o\,;\: \forall j = 1...m ;\,\forall i= 1...n.$ (Adapted from the semantic distance of the Multiplicative Prototype Model~(MPM)~\cite{minda2002comparing,zaki2003prototype} and Generalized Context Model~(GCM)~\cite{estes1986memory}).
	\label{def:pttype_distance}
\end{definition}


\theoremstyle{definition}
\begin{definition}{\textit{Distance between objects.}}
	Let $ {o_1, o_2 \in O_{c_i}}$ be a representative objects of category $c_i \in C;$ $\emph{F}_{o_1}, \emph{F}_{o_2}$ the features of objects $o_1, o_2$ respectively. We define the ~\textit{objects distance} between $o_1$ and $o_2$ as the semantic distance given by:
	\begin{eqnarray} 
	\delta(o_1,o_2) = \sum_{j=1}^{m} \left| \omega_{ij}\right|\left|f_j^1-f_j^2\right|,
	\label{eq:objects distance}
	\end{eqnarray} 
	where $\omega_{ij} \in \Omega_{i},$ $f_j^1 \in F_{o_1}\,$ and $f_j^2 \in F_{o_2} \: \forall j = 1...m ;\,\forall i= 1...n.$ (We introduce the learned weights of CNN-models in  \textit{ the psychological distance between two stimuli} defined by Medin~\cite{medin1978context}).
	\label{def:obj_distance}
\end{definition}


\theoremstyle{definition}
\begin{definition}{\textit{Feature metric space.}}
	Let ${F}_{c_i}$ be a nonempty set of all object features of the category $c_i \in C$. Since the distance function $ \delta : {F}_{c_i} \times {F}_{c_i} \to \mathbb{R}^{+}$ satisfies the axioms of \textit{non-negativity}, \textit{identity of indiscernible},	\textit{symmetry} and \textit{triangle inequality};~$\delta $ is a \textit{metric} in the features set ${F}_{c_i}$. Consequently, ~$({F}_{c_i},\delta)$ is a \textit{metric space} or \textit{feature metric space}.
	\label{def:metric_space}
\end{definition}
\vspace*{-\baselineskip}

\begin{algorithm}[ht]
	\caption{Prototype Construction}
	\label{alg:prototype}
	\begin{algorithmic}[1]
		\BState \emph{\textbf{Input}}: CNN-model $\Lambda$, objects dataset $O$, category $c_i$
		\BState \emph{\textbf{Output}}: Category Prototype ($P_i$)
		\State $ O_{c_i} \gets \left\{{o \in O: category(o) = c_i}\right\}$
		\State $ features\_block \gets \left\{ \right\}$
		\For {$ o \in O_{c_i}$ }
		\If {$o$~\textit{is\_typical}}
		\State $ F_{o} \gets \Lambda.features\_of(o)$
		\State $ features\_block \gets features\_block \cup F_{o} $
		\EndIf		
		\EndFor
		\State $ \Omega_{i},b_{i} \gets \Lambda.sofmax\_weight\_learned\_of(c_i)$
		\State $ M_{i},\Sigma_{i} \gets compute\_stats(features\_block)$
		\State \Return $ \left( {M_{i},\Sigma_{i},\Omega_{i},b_{i}}\right) $	
	\end{algorithmic}
	
\end{algorithm}

\subsection{Global Semantic Descriptor}
\label{sec:descriptor}

Our approach of object semantic description based on prototypes assumes as \textit{semantic meaning vector}, the semantic vector ($ \vec{z} = \Omega_{i}  \odot F_{o} + \bar{b_i}$) constructed from element-wise operations to compute the \textit{semantic value}~(Definition~\ref{def:semantic_value}). Furthermore, we represent the~\textit{ semantic difference vector} as the weighted \textit{residual vector}~($ \vec{r} = \left|F_{o} - M_{i}\right|$) composed of the absolute values of the difference of each object feature  with each feature of the category prototype.

Figure~\ref{fig:methodology} shows an overview of our prototype-based description model. Our \textit{Global Semantic Descriptor} model uses as requirement the prototypes priori knowledge of each CNN-model categories~(prototypes are computed using the Algorithm~\ref{alg:prototype}). After the categorization process, we use the corresponding category prototype for semantic description of object features. We show graphically in Figure~\ref{fig:methodology}d) 
how to introduce the category prototype into the global semantic description of object's features. A drawback of our representation~(Figure~\ref{fig:methodology}d)) is having high dimensionality, since it is based on the \textit{semantic meaning vector}~($\vec{z}$) and the \textit{semantic difference vector}~($\vec{\delta} = \Omega_{i} \odot \vec{r}$). The large dimensional of our feature vectors makes its practical uses unfeasible in common computer vision tasks such as semantic correspondence~\cite{han2017scnet,kim2017fcss}.

\begin{algorithm}[t]
	\caption{Global Semantic Descriptor $\psi$}
	\label{alg:global_descriptor}
	\begin{algorithmic}[1]
		\BState \emph{\textbf{Input}}: features~$F_o$, $\vec{r}$, learned weights~$(\Omega_{i},b_{i})$
		\BState \emph{\textbf{Output}}: object signature~($\psi_o$)
		\State $ \textit{meaning} \gets f\left({F_o,\Omega_{i},b_{i},\textit{meaning}}\right)$
		\State $ \textit{difference} \gets f\left({\vec{r},\Omega_{i},b_{i},other}\right)$
		\State \Return $ \textit{meaning} \oplus \textit{difference} $
	
	\end{algorithmic}
\end{algorithm}
\vspace*{-\baselineskip}

\subsubsection*{Dimensionality reduction}

Several dimensionality reduction algorithms such as PCA~\cite{abdi2010principal} and NMF~\cite{lee2001algorithms} are based on discarding features that do not generate meaningful variation. Although this approach works on some tasks, after applying these algorithms we lost the ability of data interpretation~\cite{abdi2010principal}. For the perspective of Prototypes Theory, discarding features it is no suitable when applied to the semantic space, due to the absence of necessary and sufficient definitions to categorize an object~(\textit{intensional non-discreteness}). Sometimes discarding features may mean discarding elements of the category~\cite{geeraerts2010theories}. For instance, there may be some objects within the category that do not have some of category typical features~(flying is a typical feature of bird category; however, penguin is a bird that does not fly).

We propose a simple transformation~$f(x)$ to compress our global semantic description representation of the object's features~(Figure~\ref{fig:methodology}d) in a global semantic signature~(Figure~\ref{fig:methodology}f). The final descriptor signature preserves the \textit{semantic meaning}~(Property~\ref{property:semantic_value}) and the \textit{semantic difference}~(Property~\ref{property:object_distance}) present in the first global semantic description representation. Depending on the input values, our descriptor uses  the transformation~$f(x)$ to construct global semantic signatures with different meanings within the category~(Property~\ref{property:polymorphism}).


\begin{figure} [t!]
	\begin{center}
		\includegraphics[width=1.0\linewidth]{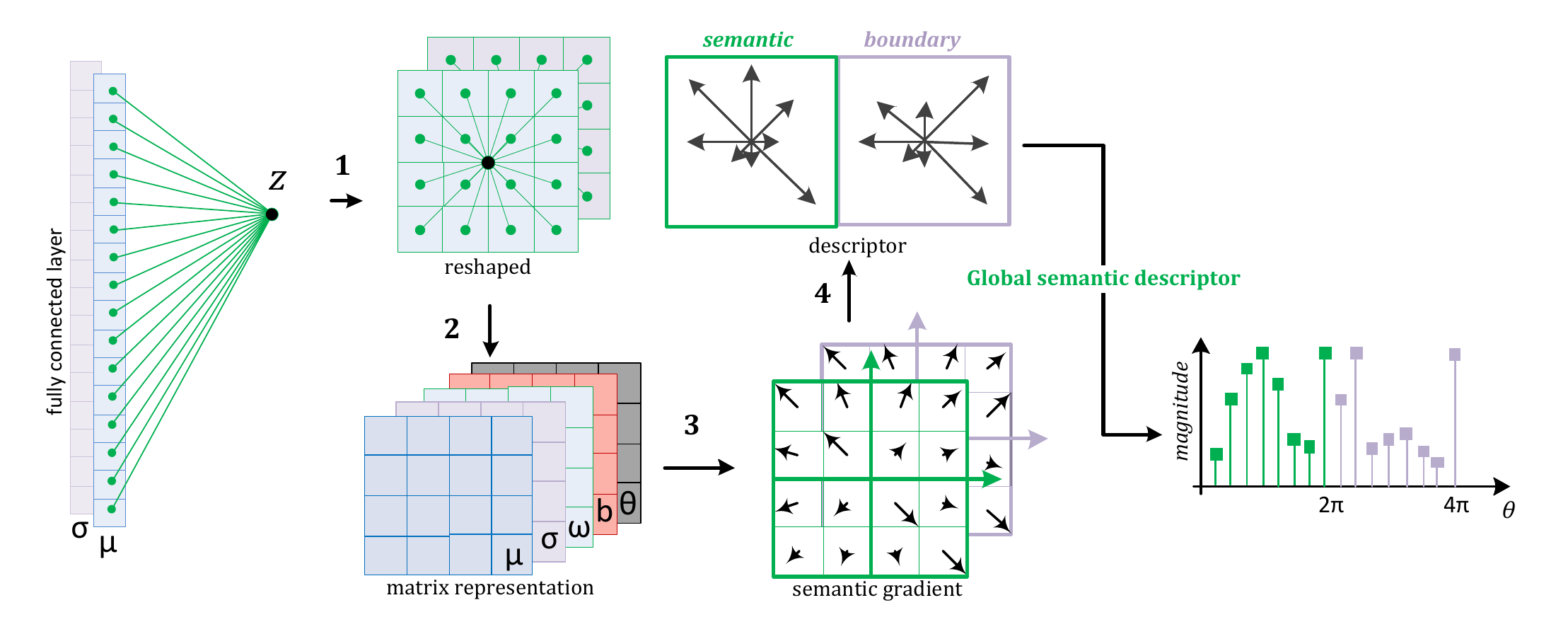}
	\end{center}
	\vspace*{-\baselineskip}
	\caption{\textit{Dimensionality reduction workflow}.
		The transformation function $f(x)$ converts the high dimensionality of semantic description representation into corresponding global semantic descriptor signature. We can see the descriptor signature computation whose taxonomy stands for the semantic meaning of $c_i$-category~(Property~\ref{property:polymorphism}\ref{property:polymorphism_category}). We also show the trivial case when the input m-dimensional vector has the same dimension as the $ \chi_{r\times r}$ auxiliary matrix~($ m=r \cdot r  \: and \: p=q=r$).}	
	\label{fig:extended}
	\vspace*{-\baselineskip}	
\end{figure}

The descriptor signature~$(\psi)$ is computed by concatenating the corresponding signatures of \textit{semantic meaning vector}~($\vec{z}$) and \textit{semantic difference vector}~($\vec{\delta}$) with our transformation $f(x)$~(see Algorithm~\ref{alg:global_descriptor}). Figure~\ref{fig:extended} shows the main steps of $f(x)$ transformation: \textbf{1)} Resizing the input vector in the best configuration of square auxiliary matrices~$\chi_{r\times r}$ and concatenate the output signatures of the flow $2$, $3$, $4$ for each $\chi_{r\times r}$; \textbf{ 2)} and \textbf{3)} constructing the \textit{semantic gradient} using the angle matrix ($\Theta_{r\times r}$) formed by the position of each feature with respect to the center of $\chi_{r\times r}$;~\textbf{4)} reducing the gradient to 8-vectors similarly to SIFT~\cite{lowe2004SIFT}.  Algorithm~\ref{alg:dim_reduction} details the steps.


\begin{algorithm}[h]
	\caption{Dimensionality Reduction $f(x)$}
	\label{alg:dim_reduction}
	\begin{algorithmic}[1]
		\BState \emph{\textbf{Input}}:
		\textit{m-dimensional vector}~$\alpha$, $\Omega_{i}, b_{i},$ \textit{type}
		\BState \emph{\textbf{Output}}: semantic signature
		\State $ \bar{b_i} \gets \frac{b_i}{m} \qquad \qquad$  { //~\footnotesize m-dimensional vector $ \bar{b_i}$ $(b_{i} = \sum_{m} \bar{b_i})$ }
		\State $ \chi_{r\times r} \gets shape(r,r) \: $ { //~\footnotesize setting source\_matrix $(\chi)$ dimension}  
		\State Finding the optimal configuration $p, q$ where $p \equiv 0\ (\textrm{mod}\ r)$, $q \equiv 0\ (\textrm{mod}\ r)$ and $ m = p \cdot q$
		\State $ \alpha, \Omega_{i},\bar{b_{i}} = reshape\_to\_matrix_{p\times q}(\alpha,\Omega_{i},\bar{b_i})$
		\State Computing angles matrix: $\Theta_{r\times r} = angles\_from( \chi_{r\times r})$
		
		\State $ signature \gets \left [  \right ]$
		\For {$ j = 1,...,\frac{p}{r}; k = 1,...,\frac{q}{r} $}
		
		\State Mapping $ \chi_{r\times r}^{jk}$ in $ \alpha, \Omega_{i},\bar{b_i} $
		\State Computing $ \vec{z_i}^{jk} $ using \textit{Hadamard product}  $\odot $.
		\State $ \vec{z_i}^{jk} = 
		\begin{cases}
		\Omega_{i}^{jk} \odot \alpha^{jk} + \bar{b_i}^{jk},&{\text{if}~\textit{type} = \textit{meaning}}\\
		\left|\Omega_{i}^{jk}\right| \odot \alpha^{jk},& \textit{otherwise}
		\end{cases}  $
		\State $g\ ^{jk} \gets vectors(\Theta_{r\times r},\left|\vec{z_i}^{jk}\right|, sign(\vec{z_i}^{jk}))$.
		
		\State $signature\ ^{jk}(l) =\sum g^{jk}(\theta), \forall \theta \in \Theta_{r\times r} : \theta_l -45 < \theta \leq \theta_l $ with $ \theta_l = l  \cdot \frac{\pi}{4}, \forall l = 1,...,8 $	
		\State $ signature \gets signature \oplus signature\ ^{jk} $
		\EndFor
		\BState \Return $signature $	
	\end{algorithmic}
\end{algorithm}

\subsubsection*{Descriptor properties}

\begin{property}{\textit{Semantic preservation.}}~The semantic descriptor signature preserves the \textit{semantic value}:~$ \int_{0}^{2\pi} \psi = \sum_{k=0}^{\left | \psi  \right |/2} \psi[k] = \hat{z}.$
	\label{property:semantic_value}  
\end{property} 
\begin{proof}
	To prove this, it suffices to follow backward through steps~$ \left [8,14  \right ]$ of Algorithm~\ref{alg:dim_reduction}. $\int_{0}^{2\pi} \psi$ = $\sum_{k=0}^{\left | \psi  \right |/2} \psi$ = $ \sum f\left({\alpha,\Omega_{i},b_{i},\textit{meaning}}\right) = \sum \sum_j \sum_k g^{jk}$= $\sum \vec{z} = \sum \Omega_{i} \odot \alpha + \bar{b_i} = \hat{z}; ~\alpha \in \left\{  {M_i, F_o}\right\}.$	
\end{proof}
%


\begin{property}{\textit{Prototypical distance preservation.}}~The object signature $\psi(o\in O_{c_i})$ preserves the \textit{prototypical distance}:~$ \int_{2\pi}^{4\pi} \psi_o = \sum_{k=\left | \psi_o  \right |/2}^{\left | \psi_o  \right |} \psi_o[k] = \delta(\emph{o},P_i).$
	\label{property:object_distance}
\end{property}
\begin{proof}
	Similar to the previous proof~(\textit{type=other}).
	$\int_{2\pi}^{4\pi} \psi =$ $\sum_{k={\left | \psi  \right |/2}}^{\left | \psi  \right |} \psi\left( {F_{o},\left|F_{o} - M_{i}\right|,\Omega_{i},b_{i}}\right)= $ $\sum f\left({\left|F_{o} - M_{i}\right|,\Omega_{i},b_{i},\textit{other}}\right)= \sum \sum_j \sum_k  g^{jk}=$ $\sum \vec{\delta} =\sum \left|\Omega_{i}\right| \odot \left|F_{o} - M_{i}\right|=$ $\delta(\emph{o},P_i).$ 
\end{proof}
\begin{property}{\textit{Structural polymorphism.}}~Our Global Semantic Descriptor
	has the polymorphic property of describing, with the same structural representation, distinctly different semantic meanings within the $c_i$-category. Consequently, our descriptor uses the category prototype $ P_i = \left({M_{i},\Sigma_{i},\Omega_{i},b_{i}}\right)$ to construct different semantic signature taxonomies:
	\begin{enumerate}[i)]
		\item an object~$o\in O_{c_i}$.~$\psi_o = \psi(F_{o},\left|F_{o} - M_{i}\right|,\Omega_{i},b_{i}) $;
		\label{property:polymorphism_object}
		\item \textit{central semantic meaning}~(abstract prototype) of $c_i$-category.~$ \psi_{P_i}=\psi(M_{i},\left|M_{i} - M_{i}\right|,\Omega_{i},b_{i}) = \psi(M_{i},\vec{0},\Omega_{i},b_{i}) $; \label{property:polymorphism_prototype}
		\item \textit{semantic meaning} of $c_i$-category.~$\psi_i=\psi(M_{i},\Sigma_{i},\Omega_{i},b_{i})$.
		\label{property:polymorphism_category}
	\end{enumerate}
	
	\label{property:polymorphism}  
\end{property} 

\section{Experiments}
\subsection{Experimental Setup}
\paragraph*{Datasets.} We conducted our experiments on two benchmark image datasets: MNIST~\cite{lecun1998} and ImageNet~\cite{ILSVRC15}. We used the $60,000$ training samples of MNIST dataset as archetype for building our prototypes. Also, we used ImageNet dataset for building our real prototypes of objects.
%
%
%

\paragraph*{Models.} We used a CNN-MNIST model  based on the LeNet architecture~\cite{lecun1998} for digit classification in the MNIST dataset. The CNN-MNIST model was used as a pilot model of our experiments as well as the VGG16 model~\cite{simonyan2014very} was the ground of our semantic description model. We used VGG16 models because its features are the basis of a variety of image processing tasks such as object detection~\cite{ren2015faster}, image annotation~\cite{murthy2015automatic}, video emotion recognition~\cite{xu2016video}, style transfer~\cite{gatys2015neural}, image alignment~\cite{han2017scnet,rocco2017convolutional}, cluster, and scene classification~\cite{lu2017jm}. Our prototype-based descriptor model is scalable and can easily be adapted to any other CNN-classification model.

\subsection{Prototype construction}
In the experiments, we computed the prototypes with CNN-MNIST and VGG16 models in MNIST and ImageNet datasets, respectively. We assume as the object features those extracted from the model layer that is at once right before the softmax layer~(see Feature Layer in Figure~\ref{fig:methodology}b). We need typical objects of the category, or any information about the typicality value (or typicality degree) of the object of a specific category, to properly build the proposed \textit{semantic prototype}. However, none of the datasets used have this information. For this reason, we used as category of typical objects only those elements that are - unequivocally - classified as category members~(Top 1) by CNN-classification models. For each category in the datasets, we extracted features of typical members and computed the \textit{semantic prototype}~(see Definition~\ref{def:semantic_pttype}) using Algorithm~\ref{alg:prototype}.


\subsection{Semantic information analysis}

\paragraph*{Prototypical behavior.} Achieving the members prototypical behavior within the category is one of the motivations and theoretical basis of our work. Nevertheless, there is no defined metric to quantify whether our representation correctly captures the category semantic meaning. This is a consequence of the fact that there is no defined metric to robustly evaluate the typicality level of an object to a category, this skill is still reserved only for human beings. 

\begin{figure}[t]
	\begin{center}
		\includegraphics[width=0.9\linewidth]{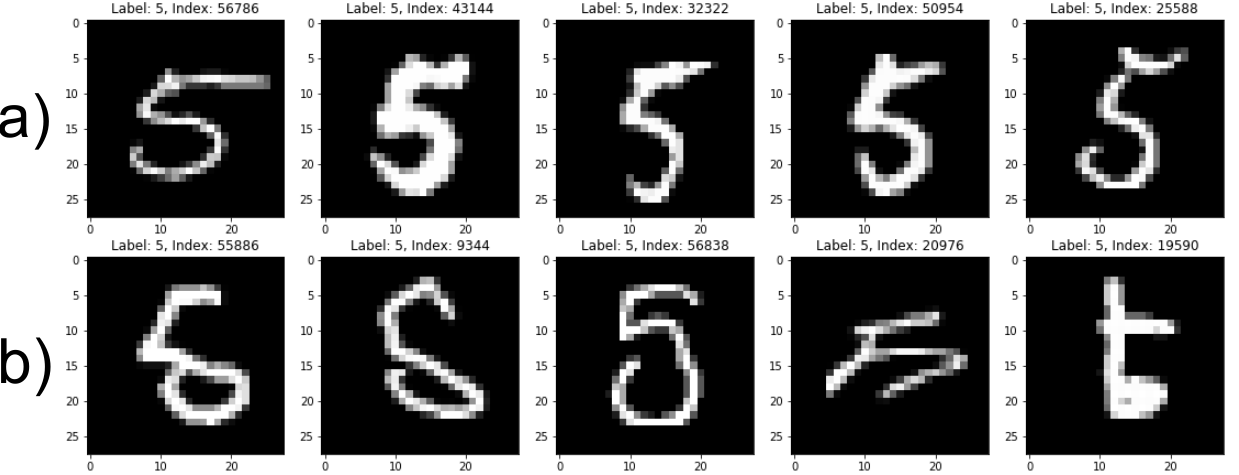}
	\end{center}
	\caption{Top-5 of the most relevant members of $c_5$-category~(number five) in the MNIST dataset.~\textit{a)}~(from left to right) Top-5 elements closest to the semantic prototype of $c_5$-category; the \emph{index} value represents the position of the object within the $c_5$-category of the dataset.~\textit{b)} Top-5 elements farthest from the semantic prototype of $c_5$-category.
	}
	\label{fig:top5_mnist}
\end{figure}

\begin{table}[t]
	\centering
	\begin{tabular}{|l|cc|cc|}
		\hline
		\multicolumn{1}{|c|}{\multirow{2}{*}{Ranking}} & \multicolumn{2}{c|}{Closest $(\delta_{min})$} & \multicolumn{2}{c|}{Farthest $(\delta_{max})$} \\ \cline{2-5} 
		\multicolumn{1}{|c|}{} & \multicolumn{1}{c}{{\small Index}} & \multicolumn{1}{c|}{{\small Value}} & \multicolumn{1}{c}{{\small Index}} & \multicolumn{1}{c|}{{\small Value}} \\ 
		\hline
		Top 1 & $56,786$ & $3.181$ & $55,886$ & $16.052$ \\
		Top 2 & $43,144$ & $3.478$ & $9,344$ & $15.888$ \\
		Top 3 & $32,322$ & $3.807$ & $56,838$ & $15.282$ \\
		Top 4 & $50,954$ & $3.896$ & $20,976$ & $14.994$ \\
		Top 5 & $25,588$ & $3.920$ & $19,590$ & $14.867$ \\
	\hline  
	\end{tabular}
	\vspace*{+2mm}
	\caption{ \textit{Prototypical distance of the $c_5$-members presented in Figure~\ref{fig:top5_mnist}.} The top-5 most relevant elements (closest and farthest) are shown based on our prototypical distance metric~$(\delta)$. It also shows the position of each element within the MNIST dataset~(index) and its semantic difference ($\delta_{value}$) with respect to the $P_5$-prototype.}
	\label{table:top5_mnist}
\end{table}

\begin{figure*}[ht]
	\begin{center}
		\includegraphics[scale=0.8,width=0.95\linewidth]{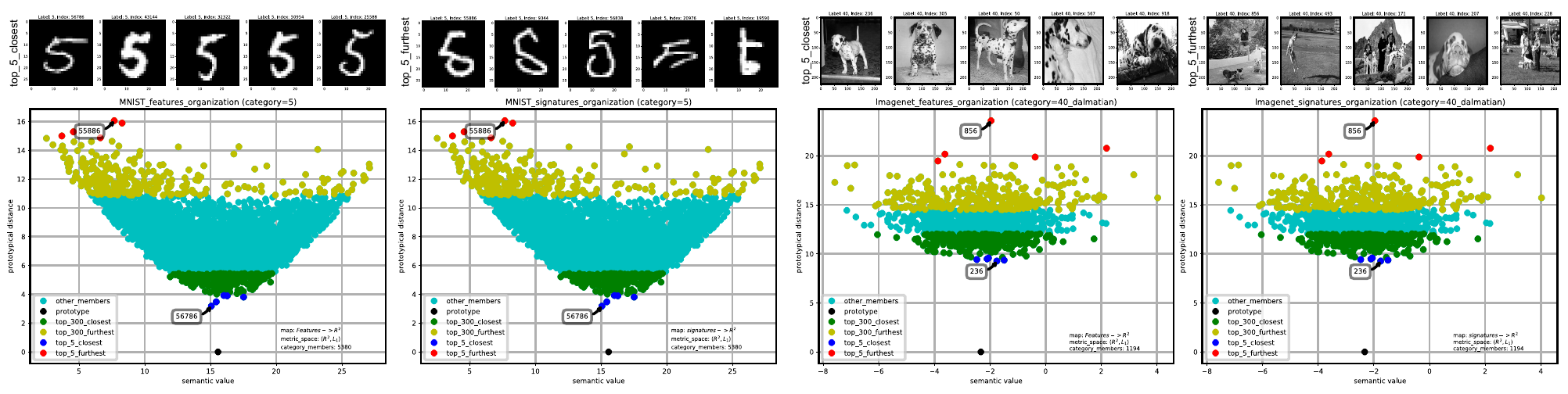}
		
	\end{center}
	\vspace*{-\baselineskip}
	\caption{\textit{Prototypical organization within categories $5$ and $40$ for MNIST and ImageNet datasets respectively.}~ 
		In the top, from left to right, the five elements closest and  furthest to the semantic prototype of each category; the index of the first element is annotated (inside the black box). Noting how the signatures domain  preserves the internal disposition of the category achieved in the feature domain.}
	\label{fig:prototypical_organization}
\end{figure*}

Our \textit{prototype model}~(\textit{semantic prototype} + \textit{prototypical distance}) tries to capture the central semantic meaning of the category. In a comparable way to the human being, we want to simulate that visually typical elements of category are organized close~(based on the prototypical distance metric) to the category prototype.

Figure~\ref{fig:top5_mnist} and Table~\ref{table:top5_mnist} present an example of the semantic meaning captured by our prototype model for members of the number five category in MNIST dataset.  As shown in Figure~\ref{fig:top5_mnist}, our proposal finds as \textit{typical} elements of number $5$ (top-5 closest) the handwritten digits with features that are, undoubtedly, distinctive of the $c_5$-category. Our model also can find the peripheral meaning of the category. Members with less representative features of the $ c_5$-category, or little readable, are placed in the periphery (top-5 farthest), away from the central meaning, but keeping the features of the category~(it still belongs to the category). Our model finds, as a human being, that it can be a $5$, but not a typical $5$.

%

Based on our experiments results~(in MNIST and ImageNet datasets), we assume that the proposed semantic prototype correctly captures the central semantic meaning of the category. Our prototypical distance has an influence on the arrangement of the elements around the category semantic prototype. Top-5 typical objects of the category are positioned close to the prototype and Top-5 less typical ones are positioned more distant from the semantic center. But, does our model  organize all category members with this prototypical organization?

\paragraph*{Prototypical organization.} Visualizing the semantic position of each category member with respect to the central semantic meaning of the category~(the abstract prototype), constitutes a simple approach to see the internal semantic structure of the entire category. The experiments in this section aim to visualize the internal semantic structure of the category using the semantic meaning encapsulated by our model for each category member. First, we need to corroborate that our prototype model can correctly interpret the object features and position it semantically within the category, keeping a prototypical organization. Second, we want to verify if the proposed semantic descriptor encodes and preserves the semantic information contained in the object features, while preserving the prototypical organization within the category.


Visualizing the category internal structure is infeasible in the m-dimensional features space since most techniques of data visualization are based on the discarding of features. From the perspective of the  Prototype Theory foundations, this approach can be problematic~(\textit{intensional non-discreteness}). For this reason, we used topology techniques to show that our model simulates the prototypical organization within the category.

Let $({F}_{c_i},\delta)$ and $(\mathbb{R}^{2},l_1)$ be the \textit{metric spaces}; and the map $\rho : {F}_{c_i} \to \mathbb{R}^{2}\mid \rho(o\in O_{c_i})= \rho(F_o)=p(\hat{z_o},\delta(o,P_i))$, where $ F_o $ are the object features, $\hat{z_o}$ is the object semantic value~(see Definition~\ref{def:semantic_value}), ~$\delta(o,P_i)$ is the prototypical distance; the point ~$p(x,y) \in \mathbb{R}^{2}$ and $l_1$ is L1-norm condition. $\rho$ maps the object to the $(\mathbb{R}^{2},l_1)$ metric space with its \textit{semantic value} and its~\textit{prototypical distance}. 

Let $o_1,o_2 \in O_{c_i}$, and $p_1 = \rho(o_1)$, $p_2 = \rho(o_2)$ the mapped point in $(\mathbb{R}^{2},l_1)$ metric space. Then, the Sum of Absolute Difference (SAD) $l_1(p_1,p_2) =l_1(\rho(o_1),\rho(o_2))=$ $\left|\hat{z_1} - \hat{z_2}\right| +\left|\delta_1 - \delta_2\right|$. Using the Definitions~\ref{def:semantic_value}, \ref{def:pttype_distance} and~\ref{def:obj_distance}; we end up with the expression:~$\delta(o_1,o_2) \leq l_1(p_1,p_2) \leq 2 \delta(o_1,o_2)$. Consequently, for every $F_{o_1}, F_{o_2} \in F_{c_i}$ and $ \varepsilon > 0 $ exists a $ \varphi = \frac{\varepsilon + 1}{2} > 0 $ such that: $\delta(o_1,o_2)<\varphi$ $ \Rightarrow$ $ l_1(\rho(o_1),\rho(o_2)) < \varepsilon$, that is,~$\rho $ \textit{is continuous}. This means that if $\rho(o_1) = p_1$,$ \forall p \in \left\{neighborhood \: of \: p_1 \right\}$ $\rho^-(p) \in \left\{neighborhood \: of \: o_1 \right\}$.

Let $({\psi}_{c_i},l_1)$ the metric space of objects descriptor signatures. Similarly, using the Properties \ref{property:semantic_value} and \ref{property:object_distance} we can show that the map $\gamma:$ $({\psi}_{c_i},l_1)$ $\to(\mathbb{R}^{2},l_1)\mid \gamma(\psi_o \in \psi_{c_i}) =p(\int_{0}^{2\pi} \psi_o,\int_{2\pi}^{4\pi} \psi_o)=p(\hat{z_o},\delta(o,P_i)) $ \textit{is continuous}. Since~$\rho$ and $\gamma$ are continuous, the behavior in $(\mathbb{R}^{2},l_1)$ metric space is equivalent to the behavior in feature metric space $({F}_{c_i},\delta)$ and descriptor's metric space~$({\psi}_{c_i},l_1)$. 

Figure~\ref{fig:prototypical_organization} shows examples of the internal semantic structure of categories mapped using $\rho$ and $\gamma$. The experiments demonstrate a prototypical organization within the category in the $(\mathbb{R}^{2},l_1)$ metric space. 
Note how the~\textit{semantic value} and \textit{prototypical distance} organize prototypically all category elements. Top5 most visually representative members of the number five in $({F}_{c_i},\delta)$ metric space~(see Figure~\ref{fig:top5_mnist}) are the same Top5 most representative in $(\mathbb{R}^{2},l_1)$ metric space. Top5 closest members are mapped~(\textit{in blue}) and positioned near the abstract prototype mapped~(\textit{in black})~(see Figure~\ref{fig:prototypical_organization}). Likewise, Top5 less representative members~(\textit{in red}) continue to be positioned in the peripheries. 
Even with different models and datasets, the internal prototypical organization of the category achieved in the descriptor signature domain (right) is identical to the prototypical organization in features domain (left). This means that our descriptor signature preserves in its taxonomy the semantic information contained in the object features.

\begin{figure}[t]
	\begin{center}
		\includegraphics[width=0.99\linewidth]{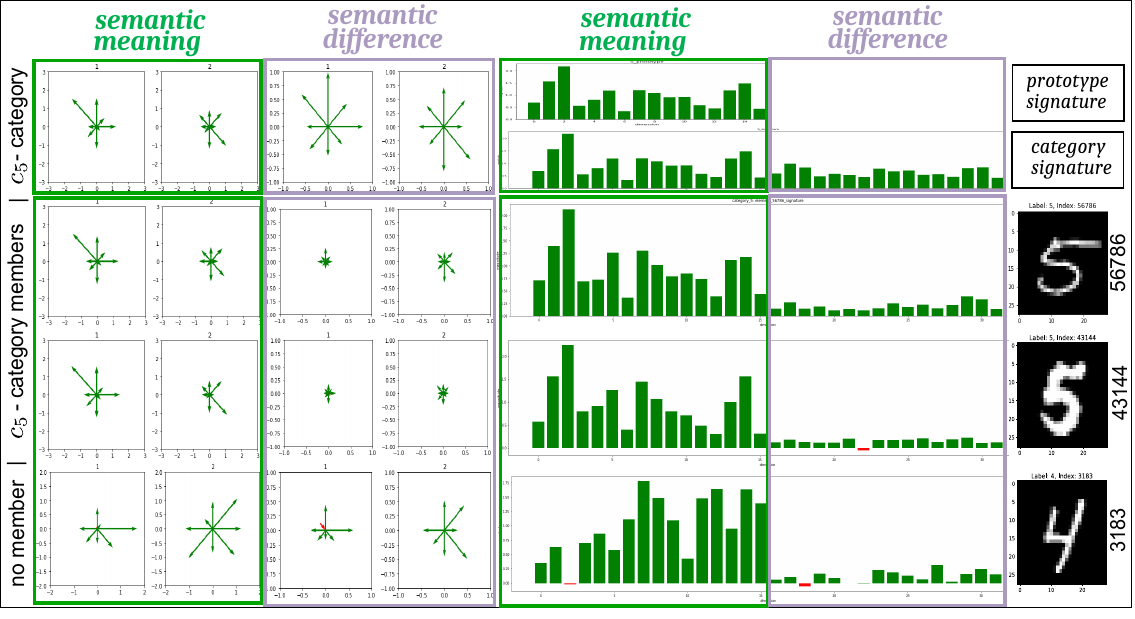}
	\end{center}
	\caption{Taxonomies of the semantic signatures constructed with our descriptor for $c_5$-category in MNIST dataset. We show the abstract prototype signature and $c_5$-category signature (\textit{semantic meaning}). In addition, we present descriptor signatures examples for two members of the $c_5$-category and a member that does not belong to the $c_5$-category. 
	}
	\label{fig:taxonomies_mnist}
\end{figure}

\paragraph*{Signature taxonomies.} Figure~\ref{fig:taxonomies_mnist} shows an example of the signatures taxonomies constructed with our descriptor using CNN-MNIST model~(signatures with size $32$). We showed the structural polymorphism property of our descriptor~(Property~\ref{property:polymorphism}) to construct signatures of the \textit{central semantic meaning}~(abstract prototype), the \textit{semantic meaning of the category} and the \textit{meaning of a category member}. The abstract prototype signature is a degenerate version of the category signature. The abstract prototype signature can be understood as the numbers distribution~(or DNA chain) that stands for the category. The category members will have a \textit{semantic meaning} with similar representation of category DNA chain. The \textit{semantic difference} of the category signature can be understood as the features boundary of all category members. Consequently, semantic information encoded in our global semantic descriptor signatures allows, easily, to recover object semantic information~(Properties~\ref{property:semantic_value}, \ref{property:object_distance}); and it also allows to interpret the object typicality within the category~(\textit{typicality score} $(\emph{o}) = 1/ \delta(\emph{o},P_i)$).

\subsection{Performance evaluation}

We evaluate the proposed semantic encoding of our Global Semantic Descriptor~(GSDP)~(version based in VGG16 model) comparing our representation against the following  image global description: GIST~\cite{oliva2001modeling}, LBP~\cite{ojala2002multiresolution}, HOG~\cite{dalal2005histograms}, Color64~\cite{li2007texture}, Color\_Hist~\cite{song2004content}, Hu\_H\_CH~\cite{haralick1973textural,hu1962visual,song2004content}, and VGG16~\cite{simonyan2014very}. Yang~\etal~\cite{yang2016joint} showed that when the features representations achieve good metrics in clustering tasks, it can generalize well when transferred to other tasks. Based in these assumptions, we evaluate our semantic encoding for verifying its usefulness and suitability in image clustering tasks.

\begin{figure}[t]
	\begin{center}
		\includegraphics[width=0.99\linewidth]{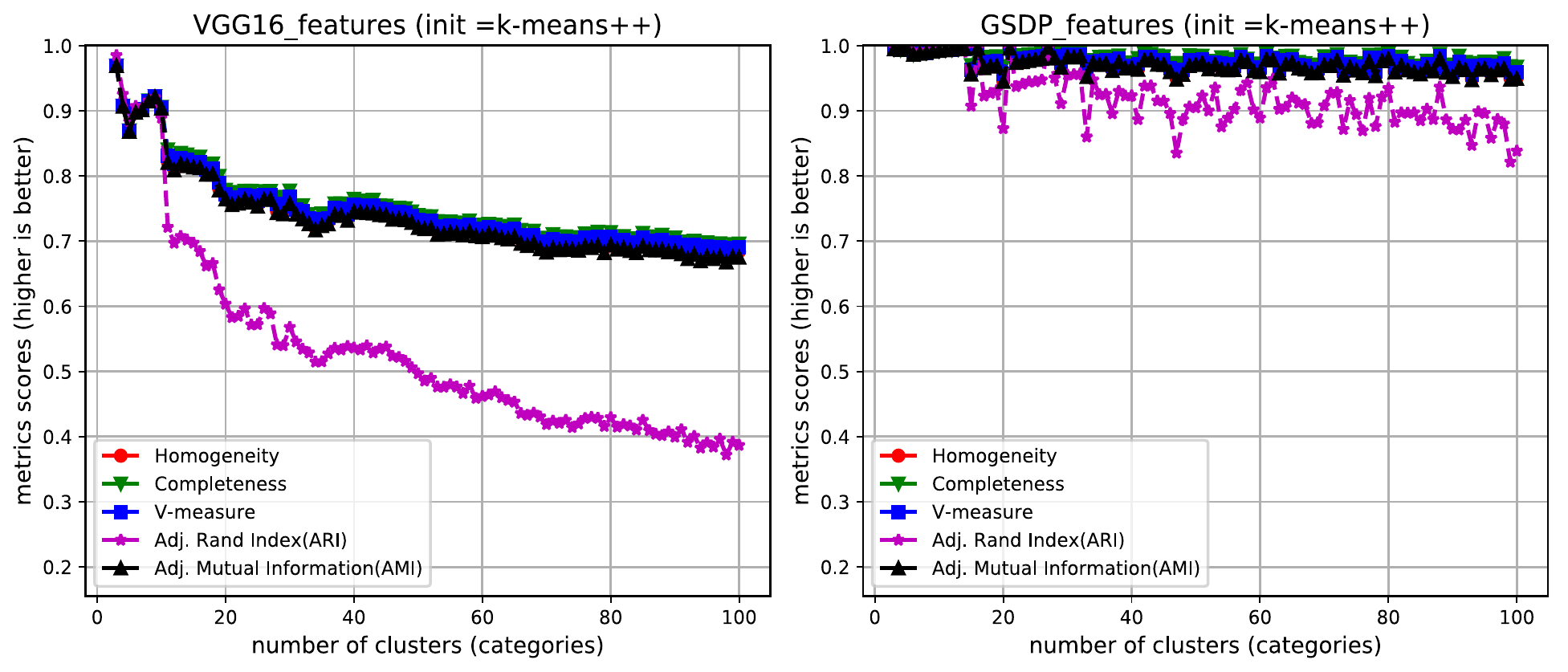}
		
	\end{center}
	\vspace*{-\baselineskip}
	\caption{Comparing the performance of VGG16 feature versus our descriptor signature  in clustering task. The Figure shows the Kmeans-metrics reached by both representations for each iterations~($100$) of our experiment.}
		
	\label{fig:Kmeans_metrics}
\end{figure}

We used the K-means algorithm for clustering $40,000$ images of the first $100$ categories of ImageNet~($400\times$ category) using the descriptors signatures. The experiment was conducted incrementally, starting with $3$ cluster (for $3$ category) and incrementing a category for each iteration. Table~\ref{table:Kmeans_metrics} shows a screenshot of K-means-metrics achieved by the selected descriptors in the first $20$ categories. Figure~\ref{fig:Kmeans_metrics} shows the Kmeans metrics behavior for VGG16 and GSDP signatures, when the number of clusters (categories) increased in each execution of algorithm.  Our GSDP descriptor keeps the semantic information of VGG16 signatures (see Figure~\ref{fig:prototypical_organization}) with a more discriminatory representation and even lower feature dimension~($256$). The results show that our descriptor encoding significantly outperforms the other image global encodings in terms of cluster metrics. The results achieved in clustering tasks encourage us to evaluate the generalization ability of our semantic representation in other computer vision tasks.

\begin{table}[t]
	\centering
    \resizebox{\columnwidth}{!}{%
	\begin{tabular}{|l|c|ccccc|}
		\hline
		\multicolumn{1}{|c|}{\multirow{2}{*}{Descriptor}} & \multirow{2}{*}{Size} & \multicolumn{5}{c|}{Metrics Scores}                                                                                   \\ \cline{3-7} 
		\multicolumn{1}{|c|}{}                            &                         & \multicolumn{1}{c|}{H} & \multicolumn{1}{c|}{C} & \multicolumn{1}{c|}{V} & \multicolumn{1}{c|}{ARI} & AMI   \\ \hline \hline
		GIST      & 960   & 0.05   & 0.05   & 0.05   & 0.01  & 0.05 \\
		LBP       & 512   & 0.02   & 0.03   & 0.03   & 0.01  & 0.02 \\
		HOG       & 1960  & 0.04   & 0.04   & 0.04   & 0.01  & 0.03 \\
		Color64   & 64    & 0.12   & 0.12   & 0.12   & 0.04  & 0.11 \\
		Color\_Hist   & 512   & 0.08   & 0.08   & 0.08   & 0.03  & 0.07 \\
		Hu\_H\_CH & 532   & 0.04   & 0.04   & 0.04   & 0.01  & 0.02 \\
		VGG16     & 4096  & 0.77   & 0.78   & 0.77   & 0.60  & 0.76 \\
		GSDP~(Our)& 256   & \textbf{0.94}   & \textbf{0.97}   & \textbf{0.95}   & \textbf{0.87}  & \textbf{0.94} \\ \hline
	\end{tabular}
}
	\vspace*{+2mm}
\caption{Kmeans cluster metrics for each evaluated descriptor. The Table shows the Kmeans-measures of clustering the first $20$ ImageNet categories~($20$ clusters): Homogeneity~(H), Completeness~(C), V-measure~(V), Adjusted Rand Index~(ARI) and Adjusted Mutual Information~(AMI).}
\label{table:Kmeans_metrics}
\end{table}

\section{Conclusions}

We introduced a novel Global Semantic Descriptor\footnote{All source code and data used will be made publicly available in our lab's website: https://www.verlab.dcc.ufmg.br/global-semantic-description/wacv2019/ } that is based on the foundations of the Prototype Theory. Our prototype-based description model does not need to be trained and it is easily adaptable to be used with any other existing CNN classification model. As shown in the experiments, our semantic descriptor is discriminative, small dimensioned, encodes the semantic information of the category, and achieves a prototypical organization of the category members. We further showed how to interpret and retrieve the object typicality information encoded in our representation. Our model proposes a starting point to introduce the theoretical foundation related to \textit{the representation of semantic meaning} and \textit{the learning of visual concepts} of the Prototype Theory in the CNN-Descriptors family.

%
%
%
%

{\small
\bibliographystyle{ieee}
\bibliography{reference}
}

\end{document}